\documentclass[10pt,twocolumn,letterpaper]{article}

\usepackage{cvpr}
\usepackage{times}
\usepackage{epsfig}
\usepackage{graphicx}
\usepackage{amsmath}
\usepackage{amssymb}
\usepackage{subfig}

\usepackage[breaklinks=true,bookmarks=false]{hyperref}

\cvprfinalcopy 

\def\cvprPaperID{****} 

\begin{document}

\title{Using Descriptive Video Services to Create a Large Data Source for Video Annotation Research}
\author{Atousa Torabi\\
Universit\'{e} de Montr\'{e}al\\
{\tt\small atousa.torabi@umontreal.ca}
\and
Christopher Pal\\
\'Ecole Polytechnique de Mont\'eral\\
{\tt\small christopher.pal@polymtl.ca}
\and
Hugo Larochelle\\
Universit\'{e} de Sherbrooke\\
{\tt\small hugo.larochelle@usherbrooke.ca}
\and
Aaron Courville\\
Universit\'{e} de Montr\'{e}al\\
{\tt\small aaron.courville@umontreal.ca}
}

\maketitle

\begin{abstract}
In this work, we introduce a dataset of video annotated with high quality natural language phrases describing the visual content in a given segment of time. Our dataset is based on the Descriptive Video Service (DVS) that is now encoded on many digital media products such as DVDs. DVS is an audio narration describing the visual elements and actions in a movie for the visually impaired. It is temporally aligned with the movie and mixed with the original movie soundtrack. We describe an automatic DVS segmentation and alignment method for movies, that enables us to scale up the collection of a DVS-derived dataset with minimal human intervention. Using this method, we have collected the largest DVS-derived dataset for video description of which we are aware. Our dataset currently includes over 84.6 hours of paired video/sentences from 92 DVDs and is growing.    
\end{abstract}

\section{Introduction}

There has been tremendous recent interest in the task of image annotation. This increased interest is likely due in part to recent innovations in the use of high capacity neural network based techniques for both images processing and for the generation of natural language phrases. Unlike many previous approaches, such techniques are able to produce both syntactically and semantically rich descriptions. However, such techniques typically require significant quantities of paired data to yield high quality results. Recent success have been fueled by the availability of large labeled datasets such as Flickr30k~\cite{Flickr} and MSCOCO~\cite{coco}. 

Video annotated in a similar way represents an even richer source of data for machine learning and artificial intelligence research. However, for video description, there have been no large datasets available. 

Descriptive Video Service (DVS) is a US-based provider of descriptive narrations that are included as audio tracks in a variety of movies and TV programs distributed on DVDs and other visual media. Their purpose is to make visual media more accessible for the visually impaired. In different regions of the world this type of annotation is known under other names such as Descriptive Video or Described Video in Canada or Audio Description in the UK. Lists of DVDs with DVS audio tracks are available on many websites such as that of the WGBH media group ~\cite{WGBH}. In Canada, the Canadian Radio-television and Telecommunications Commission (CRTC) maintains a list of television channels with described video ~\cite{CRTC}. While the WGBH television station was an early pioneer in the creation of this technology, there are now a number of providers of similar services. In our work here we outline our solution to the semi-automated construction of a large dataset using DVDs with DVS audio narrations. We also present the Montreal Video Annotation Dataset (M-VAD) created using this technique. 

DVS narrations describes key visual elements of the video such as changes in the scene, people's appearance, gestures, actions, and their interaction with each other and the scene's objects in concise but precise language. DVS soundtracks on DVDs are carefully positioned within movies to fit in natural pauses in the dialogue and are mixed with the original movie soundtrack by professional post-production. Video description is written by professional writers using rich natural sentences and their maximum misalignment with the described scene is limited to 2 seconds. For these reasons the use of DVS or other similar audio narrations is therefore appealing in terms of the quality of the narrations and the relatively high degree of alignment. Recently and in parallel to our work, it was shown that DVS on DVDs are a better source of visual description ~\cite{Rohrbach2015} in that they are more precisely aligned with the visual content of a movie compared to movie scripts. Movie scripts are written in advance before movie production and are sometimes changed during movie production. Moreover, many automatic alignment techniques for scripts and movies are imprecise and require heavy manual cleaning if they are to be used as a source of high quality training and evaluation data. As such, the automated use of scripts to construct large sources of paired video and annotations is problematic.

Despite the advantages offered by DVS, creating a completely automated approach for extracting the relevant narration or annotation from the audio track and refining the alignment of the annotation with the scene still poses some challenges. In this paper we discuss our solution. 

One of the main challenges in automating the construction of a video annotation dataset derived from DVS audio is accurately segmenting the DVS output, which is mixed with the original movie soundtrack. In \cite{Rohrbach2015}, DVS segments are detected by exploiting a similarity comparison between the DVS track and the original movie soundtracks, using Fast Fourier Transform (FFT) techniques. While comparisons using similarity measures of this sort provide a partial solution to the alignment problem, we are interested in automating the annotation and video alignment as much as possible.


In this work we describe our methods for DVS narration isolation and video alignment. DVS narrations are typically carefully placed within key locations of a movie and edited by a post-production supervisor for continuity. For example, when a scene changes rapidly, the narrator will speak multiple sentences without pauses. Such content should be kept together when describing that part of movie. If a scene changes slowly, the narrator will instead describe the scene in one sentence, then pause for a moment, and later continue the description. By detecting those short pauses, we are able to align a movie with video descriptions automatically.

We have collected a new DVS-derived dataset from 92 DVDs \footnote{List of DVDs in our collection:}
Other researchers may contact us to discuss how they may obtain access to the information necessary to replicate a well defined evaluation that follows the experiments with this dataset presented \cite{Li}. We present some qualitative examples of automatically generated descriptions from our work in \cite{Li} in Section 5 of this paper.


%
The DVDs used to construct our dataset can be bought from the Amazon~\cite{amazon} or HMV~\cite{hmv} websites.  

We believe the M-VAD dataset will be useful for a wide range of different research areas, especially for high capacity deep learning models for "in the wild" video description. 
\footnotetext{21 JUMP STREET, 30 MINUTES OR LESS, 40 YEAR OLD VIRGIN, 500 DAYS OF SUMMER, ABRAHAM LINCOLN VAMPIRE HUNTER, A GOOD DAY TO DIE HARD, A THOUSAND WORDS, BAD TEACHER, BATTLE LOS ANGELES, BIG MOMMAS LIKE FATHER LIKE SON, BLIND DATING, BRUNO, BURLESQUE, CAPTAIN AMERICA, CHARLIE ST CLOUD, CHASING MAVERICKS, CHRONICLE, CINDERELLA MAN, COLOMBIANA, DEAR JOHN, DEATH AT A FUNERAL, DINNER FOR SCHMUCKS, DISTRICT 9, EASY A, FLIGHT, FRIENDS WITH BENEFITS, GET HIM TO THE GREEK, GHOST RIDER SPIRIT OF VENGEANCE, GREEN ZONE, GROWN UPS, HANSEL GRETEL WITCH HUNTERS, HOW DO YOU KNOW, HUGO, IDES OF MARCH, INSIDE MAN, IN TIME, IRON MAN2, ITS COMPLICATED, JACK AND JILL, JULIE AND JULIA, JUST GO WITH IT, KARATE KID, KATY PERRY PART OF ME, KNOCKED UP, LAND OF THE LOST, LARRY CROWNE, LIFE OF PI, LITTLE FOCKERS, MORNING GLORY, MR POPPERS PENGUINS, NANNY MCPHEE RETURNS, NO STRINGS ATTACHED, PARENTAL GUIDANCE, PERCY JACKSON LIGHTENING THIEF, PROMETHEUS, PUBLIC ENEMIES, ROBIN HOOD, RUBY SPARKS, SALT, SANCTUM, SNOW FLOWER, SORCERERS APPRENTICE, SOUL SURFER, SPARKLE 2012, SUPER 8, THE ADVENTURES OF TINTIN, THE ART OF GETTING BY, THE BIG YEAR, THE BOUNTY HUNTER, THE CALL, THE DESCENDANTS, THE GIRL WITH THE DRAGON TATTOO, THE GUILT TRIP, THE ROOMMATE, THE SITTER, THE SOCIAL NETWORK, THE VOW, THE WATCH, THINK LIKE A MAN, THIS MEANS WAR, THOR, TITANIC1, TITANIC2, TOOTH FAIRY, TRUE GRIT, UGLY TRUTH, WE BOUGHT A ZOO, WHATS YOUR NUMBER, XMEN FIRST CLASS, YOUNG ADULT, ZOMBIELAND, and ZOOKEEPER.}
\begin{figure*}[t]
\center
\includegraphics[width=17cm]{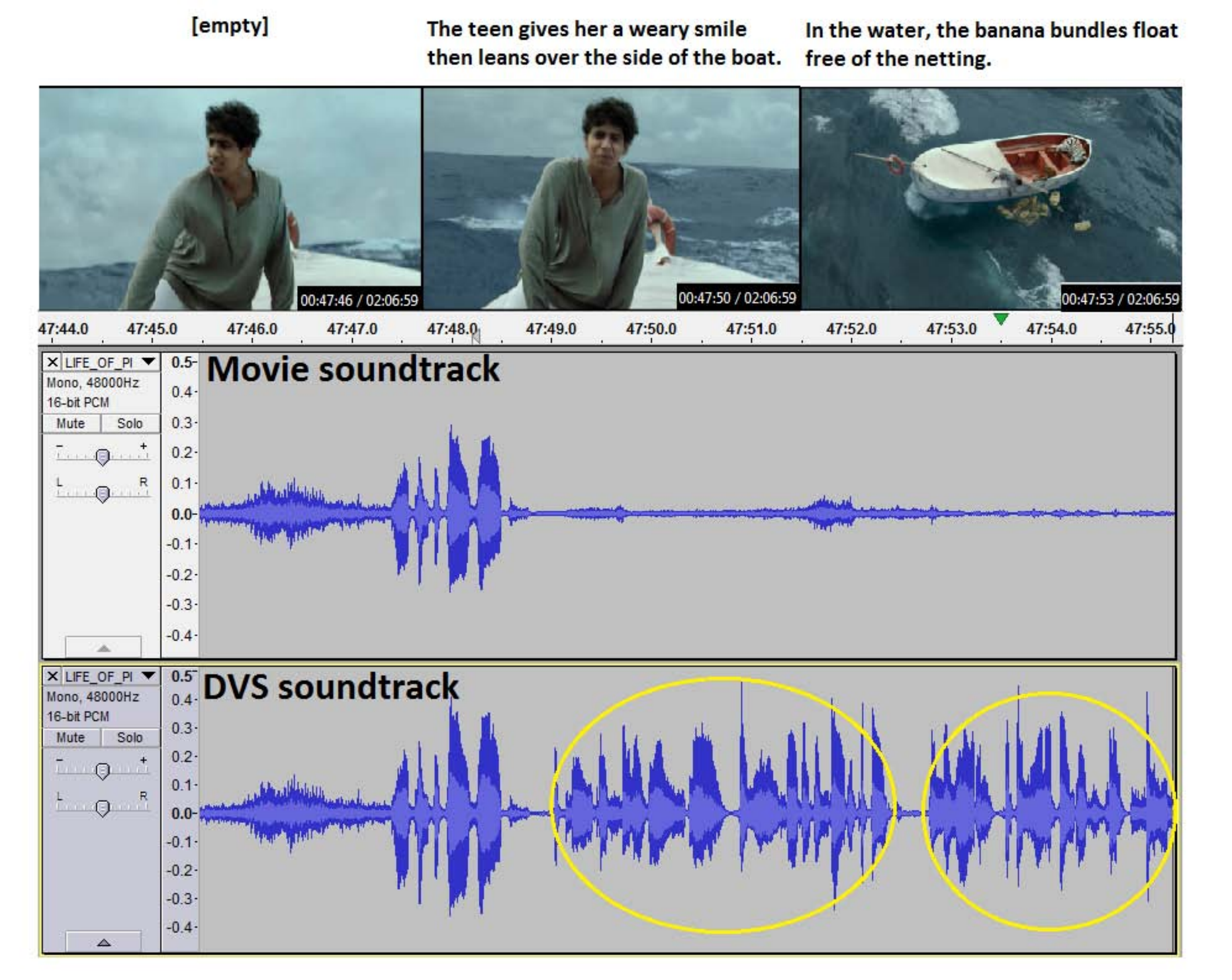}            
\caption{DVS dataset collection. From the movie "Life of the Pie". Line 2 and 3: Vocal isolation of movie and DVS soundtrack. Second and third rows shows movie and DVS audio signals after voice isolation. The two circles show the DVS segments on the DVS mono channel track. A pause (flat signal) between two DVS narration parts shows the natural DVS narration segmentation while the narrator stops and then continues describing the movie. We automatically segment DVS audio based on these natural pauses. At first row, you can also see the transcription related to first and second DVS narration parts on top of second and third image shots. }
\label{fig:thirdlayers}
\end{figure*}

\begin{table*}[ht]
\caption{M-VAD dataset overal statistics.}
\label{tab1}
\centering
\begin{tabular}{|l|c|c|c|c|c|c|}
\hline
&Movies& \# Words & \# Paragraph & \# Sentences & Avg. video length & Total video length \\     
\hline
 Un-Filtered     & 92 & 531,778  & 52,683 & 59,415 & 6.3 sec. & 91.0 h.\\
 Filtered  	& 92 & 510,933	& 48,986 & 55,904 & 6.2 sec. & 84.6 h.\\
\hline
\end{tabular}
\end{table*}  

\section{Related Work}
Recently, there have been important advances in deep learning for natural language description generation from images~\cite{Kiros2014,donahue2014long,KarpathyCVPR14,vinyals2014show}. Existing, relatively big open-domain image datasets such as Filckr8k~\cite{flickr8k}, Filckr30k~\cite{Flickr}, MSCOCO~\cite{coco}, SBU~\cite{SBUs} have played an important role in recent breakthroughs in image annotation with natural sentences. In contrast, video annotation with natural language descriptions has not been investigated as extensively and with as much success. Recently, \cite{VenugopalanXDRMS14} proposed to use a 2D-ConvNet and an RNN for video description generation by transferring knowledge from 100,000 images with captions. One of the main issue associated to automatic video annotation is the lack of open-domain, paired video/caption datasets. The majority of available video datasets are toy domains with with small word vocabularies such as TaCos~\cite{rohrbach13iccv}, TaCos Multi-Level~\cite{tacoMulti}, YouCook~\cite{youtube}, and MSVD~\cite{chen11}. Recently, work parallel to ours presented a movie description dataset from DVDs, where descriptions come either from DVS from scripts. In their comparative study, they have shown that DVS is a richer and more accurate source of descriptions, while the movie scripts are less reliable in many cases and are not as well temporally aligned with movies~\cite{Rohrbach2015}. In their work, a semi-automatic FFT-based DVS segmentation method is used, that requires human alignment in post-processing. In their collection of 72 movies, the description of 46 of these movies (i.e. 34.7 hours) came from DVS and the rest came from scripts. To the best of our knowledge, our dataset is the largest DVS-derived movie dataset available. We are currently working on further increasing its size.
\begin{figure*}[t]
\center
\subfloat(a){\includegraphics[width=17cm]{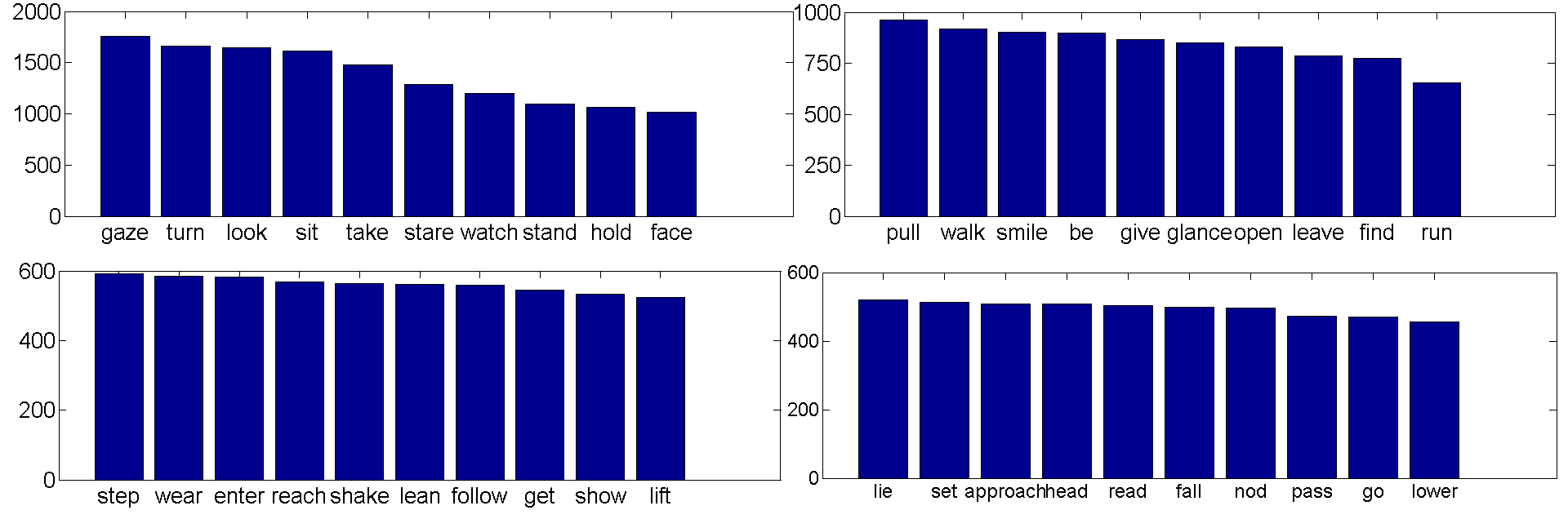}} 
\subfloat(b){\includegraphics[width=17cm]{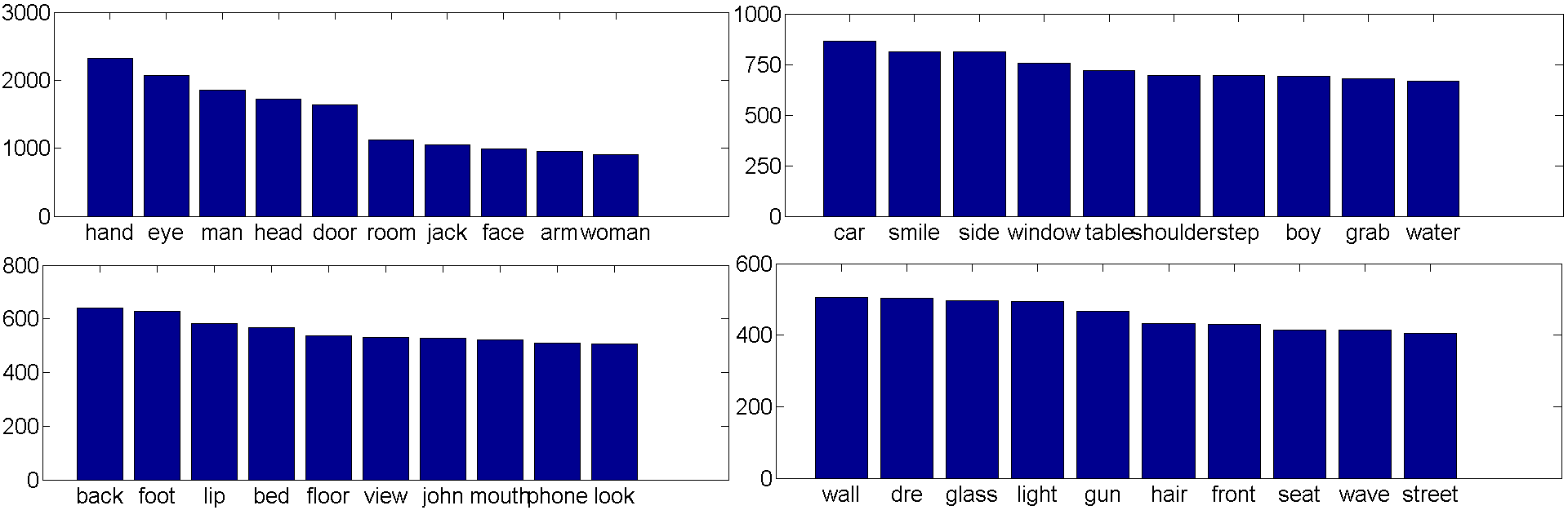}} 
\caption{(a) 40 most frequent verbs and (b) 40 most frequent nouns.}
\label{fig:verbnounstat}
\end{figure*}

\begin{table*}[ht]
\caption{Comparing video annotation datasets.}
\label{tab2}
\centering
\begin{tabular}{|l|c|c|c|c|c|}
\hline
Dataset & Context & Videos & Description source & \# Clips & \# Sentences  \\       
\hline
TACoS ~\cite{rohrbach13iccv}  & cooking  & 123 & crowd  & 7,206 & 18,227 \\
TACoS multilevel ~\cite{tacoMulti}  & cooking	& 273 & crowd	& 14,105 & 52,593\\
MSVD ~\cite{chen11}  & various & - & crowd	& 1,970 & 70,028\\
movie description ~\cite{Rohrbach2015}  & various & 72 & Script + DVS & 54,076 & 54,076\\
movie description (DVS part) ~\cite{Rohrbach2015} & various & 46 & DVS & 30,680 & 30,680\\ 
M-VAD (ours)  &various & 92 & DVS	& 48,986 & 55,904\\
\hline
\end{tabular}
\end{table*} 

\section{DVS-Derived Dataset Collection}
DVS narrations audio track on DVD is a rich source of data that describes videos in professionally written natural sentences. According to the American Council of the Blind (ACB)~\cite{blind} and the WGBH media access group  ~\cite{WGBH}, the number of DVDs featured with DVS is growing rapidly every year. This makes DVS narrations an appealing source of data for collecting large paired video/sentence datasets. The DVS narrations are mixed with original movie sound track. We've found that efficient methods with minimal human intervention were sufficient to build a large DVS-derived dataset by processing the large amount of DVS audio, aligning it with the movies and transcribing it to text. In the subsequent sections, we describe the steps we followed for our data collection in details.   

\subsection{DVS Narrations Segmentation Using Vocal Isolation}
Vocal isolation technique boosts vocals including dialogues and DVS narrations while suppressing background movie sound in stereo signals. This technique is used widely in karaoke machines for stereo signals to remove the vocal track by reversing the phase of one channel to cancel out any signal perceived to come from the center while leaving the signals that are perceived as coming from the left or the right. The main reason that we use vocal isolation for DVS segmentation is based on the fact that DVS narration is mixed within natural pauses in the dialogue, when there is DVS narration there is no dialogue. Therefore, in vocal isolated signals when the narrator speaks, the movie signal is almost a flat line relative to DVS signal allowing us to cleanly separate the narration from other dialogue by comparing two signals. Figure~\ref{fig:thirdlayers} illustrates an example from movie "Life of Pi" where in the original movie soundtrack there are the sounds of ocean waves in the background.

Our approach has three main steps: first we isolate vocal including dialogues and DVS narrations, second we separate the DVS narrations from dialogues, and finally we apply a simple thresholding method to extract DVS segment audio tracks.  

We isolate vocals using Adobe Audition's center channel extractor~\cite{audition} implementation to boost DVS narrations and movie dialogues while suppressing movie background sounds on both DVS and movie audio signals. We align the movie and DVS audio signals by taking a fast Fourier transform (FFT) of two audio signals, computing cross-correlation, a measure of similarity vs offset and select the offset which corresponds to peak cross-correlation. After alignment, we apply Least Mean Square (LMS) noise cancellation and subtract the DVS mono squared signal from the movie mono squared signal in order to suppress dialogue in the DVS signal. For the majority of movies, applying LMS results in cleaned DVS narrations for DVS audio signal. Even in cases where the standard movie audio signal and DVS signal shapes are very different - due to the DVS mixing process - our procedure is sufficient for the automatic segmentation of DVS narration.

Finally, we extract the DVS audio tracks by detecting the beginning and end of DVS narration segments in the DVS audio signal (i.e. where the narrator starts and stops speaking) using a simple thresholding method that we applied to all DVDs without changing the threshold value. In contrast, in recent work~\cite{Rohrbach2015}, DVS audio segementation was done by comparing the similarity of the two signals (the standard movie soundtrack and the soundtrack+DVS) in the Fourier domain against a threshold. We have found that with a similar approach, these types of thresholds usually have to be adjusted for each movie. Moreover we have found that such thresholds depend on the intensity of the background audio signal and the level of the DVS narrator's voice. Segmenting the DVS signal without any background audio suppression requires human cleaning.

\subsection{Movie/DVS Alignment And Professional Transcription}
DVS audio narration segments are time-stamped based on our automatic DVS narration segmentation. In order to compensate potential 1-2 seconds misalignment between the DVS narrator speaking and the corresponding scene in the movie, we automatically added two seconds to the end of each video clips, without any human intervention. We believe that in ~\cite{Rohrbach2015}, movie/DVS narrations alignment has been done manually partially because of some imprecise detection of the DVS narrations segments in DVS signal as we explained above. 

In order to obtain high quality text descriptions, the DVS audio segments were transcribed with more than 98 percent transcription accuracy, using professional transcription services~\cite{transInc}. These services use a combination of automatic speech recognition techniques and human transcription to produce a high quality transcription. Our audio narration isolation technique allows us to process the audio into small, well defined time segments and reduce the overall transcription effort and cost.

\begin{table*}[ht]
\caption{Corpus POS statistics. }
\label{tab3}
\centering
\begin{tabular}{|l|c|c|c|c|}
\hline
\ Total vocabulary  & \# Nouns & \# Verbs & \# Adjective & \# Adverb\\       
\hline
 17,609 & 9,512	& 2,571 & 3,560 & 857\\
\hline
\end{tabular}
\end{table*}  

\begin{figure*}
\begin{center}
\includegraphics[width=17 cm]{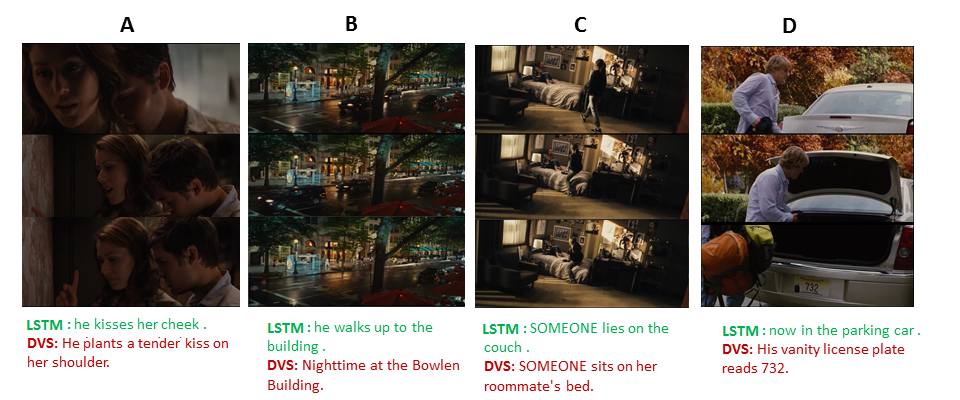} 
\end{center}
\caption{Four samples of generated sentences by our LSTM model and DVS narrations from the movies: A: Charile st. cloud, B: How do you know, C: The roommate, and D: The big year.}
\label{fig:LSTM}
\end{figure*}

\section{DVS-Derived Dataset Statistics and Comparison With Other Datasets}
Our dataset currently contains video clips from 92 DVDs, covering a variety of movie genre. Since in the beginning and end of the movie, the DVS narrator reads any text shown in the movie (e.g. movie credits), we discard sentences that are time-stamped within three minutes at the beginning and 4 minutes at end of the movie. Table ~\ref{tab1} shows the overall statistics for both our un-filtered and filtered dataset. The filtered data contains 48,986 video clips, with an average length of 6.2 seconds (for a total of 84.6 h). The number of sentences (counted based on the number of periods in the dataset) is 55,904, meaning that some clips are paired with more than one sentence. 

In table~\ref{tab2}, we compare our dataset with existing paired sentence/video datasets. To the best of our knowledge, our dataset is the biggest DVS-derived dataset available. The recent movie description dataset of \cite{Rohrbach2015} is the most similar, however the source for the text descriptions is a combination of movie scripts and DVS. The DVS part consists in only 46 DVDs, while our dataset contains 92 DVDs. Other existing datasets have a limited number of videos and domains, where the source of the text description is based on crowdsourcing. In contrast, our DVS-derived dataset has a variety of videos with professionally written captions.

We tagged all words in the dataset corpus using the Standford Part-Of-Speech (POS) tagger toolbox~\cite{pos}. Table~\ref{tab3} illustrates the vocabulary size, number of nouns, verbs, adjective, and adverbs in the DVS dataset. It is interesting that the number of adjectives is larger than the number of verbs, which shows that the DVS is describing the characteristics of visual elements in the movie in some detail. Also, figure~\ref{fig:verbnounstat} shows the 40 most frequent verbs and nouns in our DVS dataset. It is interesting that among the 10 most frequent verbs, 5 of them are synonyms of "seeing" (i.e. gaze, look, stare, watch, and face). 

\section{DVS-Derived Dataset Corpus Preparation}
In our DVS dataset there is over 500 proper names. When training a video description model, we are usually not interested in learning how to generate such names. Moreover, our recent work on training an LSTM-based model on this dataset suggests that removing proper names from the dataset is beneficial. In this dataset We have replaced all people's name with a single token word (e.g.\ "SOMEONE").

We also provide an official training/validation/test split for our dataset, consisting of 38,949, 4888 and 5149 video clips respectively. This split balances DVD genres within each set, which is motivated by the fact that the vocabulary used to describe, say, an action movie could be very different from the vocabulary used in a comedy movie. 

Figure~\ref{fig:LSTM} shows some of the preliminary qualitative results of our recent LSTM video description model~\cite{Li} on this dataset, based on our official dataset split. "LSTM" is the generated sentence and "DVS" is the original DVS sentence. In these examples, even though the samples are not necessarily the same as the DVS reference,  they are still meaningful and closely related to the visual context.

\section{Conclusion}
In this work, we have introduced a new, large DVS-derived video dataset. It is publicly available for the research community. Our automatic DVS segmentation and alignment method enabled us to collect this dataset with minimal human intervention. We also provide a balanced split of data for defining an official task.

{\small
\bibliographystyle{ieee}
\bibliography{egbib}
}

\end{document}